\documentclass{article}
\usepackage{ijcai24}

\usepackage{times}
\usepackage{soul}
\usepackage{url}
\usepackage[hidelinks]{hyperref}
\usepackage[utf8]{inputenc}
\usepackage[small]{caption}
\usepackage{graphicx}
\usepackage{amsmath}
\usepackage{amsthm}
\usepackage{booktabs}
\usepackage{algorithm}
\usepackage{algpseudocode}
\usepackage[switch]{lineno}
\usepackage{amsmath,amssymb,diagbox,tabularx,svg,subcaption,multirow,setspace,titlesec}
\usepackage{algpseudocode}

\title{Weakly Supervised Change Detection via Knowledge Distillation and Multiscale Sigmoid Inference}

\author{
    Binghao Lu, Caiwen Ding, Jinbo Bi, Dongjin Song\footnote{Corresponding author: Dongjin Song}
    \affiliations
    Department of Computer Science and Engineering, University of Connecticut
    \emails
    binghao.lu, caiwen.ding, jinbo.bi, dongjin.song@uconn.edu
}

\begin{document}

\maketitle

\begin{abstract}

Change detection, which aims to detect spatial changes from a pair of multi-temporal images due to natural or man-made causes, has been widely applied in remote sensing, disaster management, urban management, \textit{etc.} Most existing change detection approaches, however, are fully supervised and require labor-intensive pixel-level labels. To address this, we develop a novel weakly supervised change detection technique via Knowledge Distillation and Multiscale Sigmoid Inference (KD-MSI) that leverages image-level labels. In our approach, the Class Activation Maps (CAM) are utilized not only to derive a change probability map but also to serve as a foundation for the knowledge distillation process. This is done through a joint training strategy of the teacher and student networks, enabling the student network to highlight potential change areas more accurately than teacher network based on image-level labels. Moreover, we designed a Multiscale Sigmoid Inference (MSI) module as a post processing step to further refine the change probability map from the trained student network. Empirical results on three public datasets, \textit{i.e.}, WHU-CD, DSIFN-CD, and LEVIR-CD, demonstrate that our proposed technique, with its integrated training strategy, significantly outperforms the state-of-the-art. Code is available at \href{https://github.com/BinghaoLu/KD-MSI}{https://github.com/BinghaoLu/KD-MSI}.
\end{abstract}

\section{Introduction}

\begin{figure}
\begin{center}
   \includegraphics[width=0.9\linewidth]{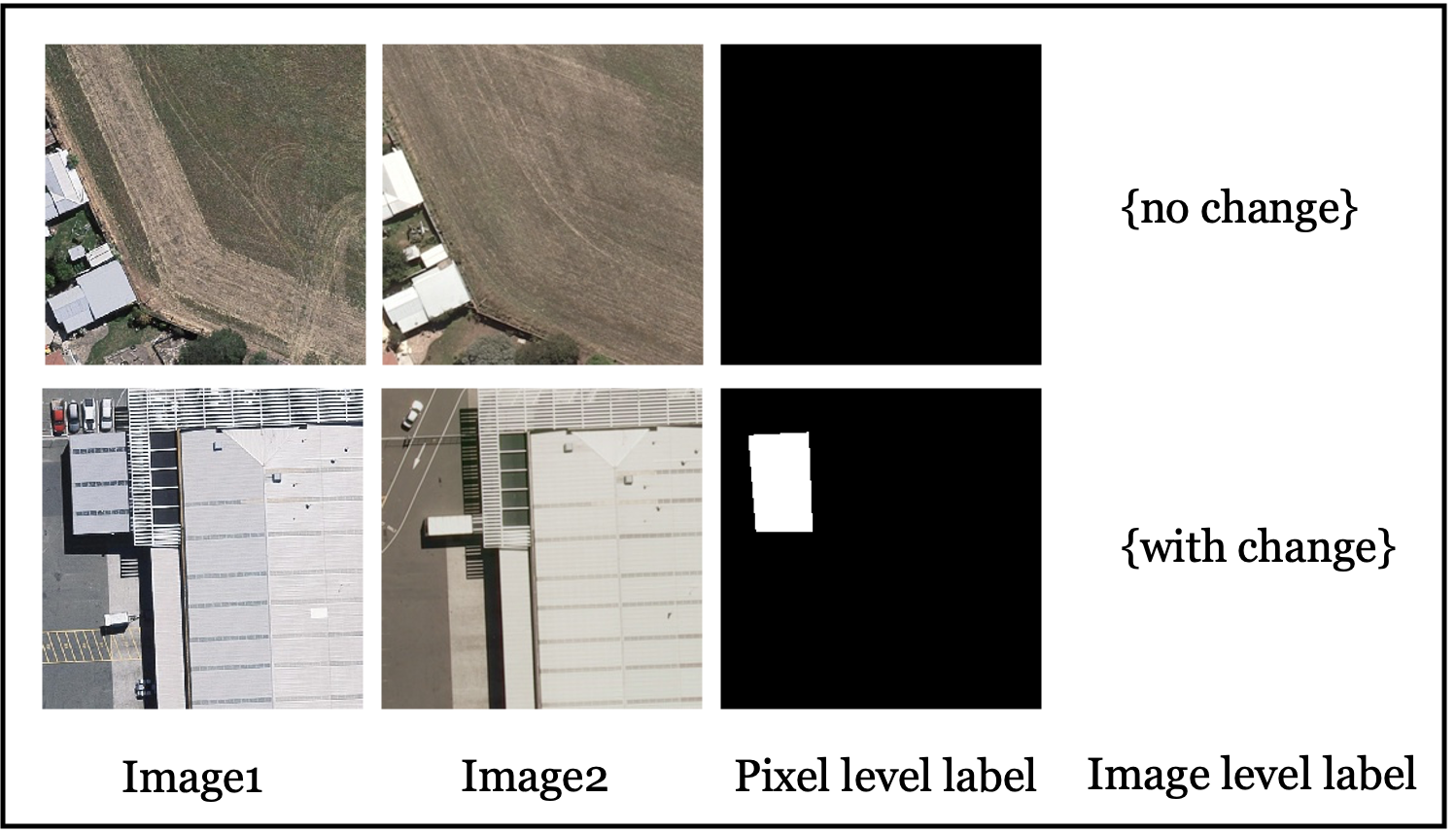}
\end{center}
   \caption{\small Comparison of change detection methods with pixel-level supervision (third column) vs. image-level supervision (fourth column).}
\label{fig:1}
\end{figure}

Change detection aims to identify the changes of objects within the same geological location across different periods. It has been applied in various real-world applications, \textit{e.g.}, disaster management, urban planning, visual surveillance, and resource management~\cite{bouziani2010automatic,goyette2012changedetection,jiang2022survey,sublime2019automatic,lee2021local,ye2021near}. 
Recently, tremendous progress has been made in deep learning-based change detection tasks as deep neural networks show their superiority in producing more effective representations\mbox{~\cite{jiang2022survey}}. Most existing approaches, however, focus on fully supervised change detection tasks and require a massive amount of pixel-level labels. This is, however, time-consuming and labor-intensive~\cite {jiang2020pga}. An alternative is to adopt image-level labels and leverage weakly supervised learning to train the change detection model. A comparison of change detection tasks with pixel-level labels versus image-level labels is depicted in Figure \ref{fig:1}.

Although it is challenging to perform weakly supervised change detection, several attempts have been made to tackle this problem in the past few years~\cite{khan2016learning,andermatt2020weakly,kalita2021land,wu2023fully,huang2023background}. For instance, Khan et al.~\cite{khan2016learning} and Andermatt et al.~\cite{andermatt2020weakly} developed conditional random fields with image-level labels to perform change detection. Kalita et al.~\cite{kalita2021land} employed principal component analysis together with K-means clustering to produce the change map based on a Siamese neural network architecture. Wu et al.~\cite{wu2023fully} developed a fully convolutional change detection framework based on a generative adversarial
network to facilitate weakly supervised change detection. Huang et al.~\cite{huang2023background} developed a new augmentation method to mix the background region of the image pair to enhance weakly change detection performance. Despite the progress, there is still a large gap between the coarse image-level labels and satisfactory fine-grained pixel-level change detection results for real-world applications.


To bridge the aforementioned research gap, we develop a novel weakly supervised change detection technique by leveraging a knowledge distillation framework and a multiscale sigmoid inference module. Specifically, we adapt Class Activation Maps (CAM)~\cite{zhou2016learning}, a technique that is originally used to visualize which regions contribute the most to the prediction made by a convolutional neural network, to incorporate a pair of input images via Siamese neural networks (teacher network) to highlight the potential change area based on image-level labels, \textit{i.e.}, change or no change. Although CAM can capture the discriminative region of potential changes, the results are still relatively coarse and may contain incorrectly labeled pixels. To refine this, we implement a joint training strategy where the student network's change probability map is trained under the guidance of the teacher network through the knowledge distillation process. This approach allows the student network to learn the nuanced patterns and knowledge from the teacher network's CAM, enhancing its ability to generate a more precise change probability map. Furthermore, we design a multiscale sigmoid inference module as a post processing step to further enhance the change probability map of the student network. Extensive experiments demonstrate that our proposed model, with its structured joint training approach and the integration of the multiscale sigmoid inference module, significantly outperforms state-of-the-art over three publicly available datasets, \textit{i.e.}, LEVIR-CD, WHU-CD, and DSIFN-CD.

\section{Related Work}
Change detection and semantic segmentation are two fundamental tasks in the field of remote sensing and computer vision. 
Our proposed method is closely related to fully supervised change detection, weakly supervised semantic segmentation and weakly supervised change detection.


\subsection{Fully-supervised change detection}
Fully supervised change detection relies on pixel level labeled data to train the model~\cite{shi2020change}~\cite{khelifi2020deep}. Deep learning based fully supervised change detection methods has been popular since last decades. In the beginning, convolutional neural network based methods are dominant, recently the transformer based methods entered the stage. Daudt et al.~\cite{daudt2018fully} designed the first end-to-end UNet structure for fully supervised change detection with their designed FC-EE, FC-Siam-conc, and FC-siam-diff modules. There are also many convolutional neural netowrk based change detection methods use VGG ~\cite{simonyan2014very} or Resnet ~\cite{he2016deep,zheng2021building} based backbones. For example, IFNet~\cite{zhang2020deeply} and DTCDSCN~\cite{liu2020building} used VGG16 and ResNet34 as backbone respectively to do change detection. As for the transformer based change detection, SwinSuNet~\cite{zhang2022swinsunet} adopted swin transformer as their backbone and BIT ~\cite{chen2021remote} adpots ResNet18 and Vision Transformer as backbone for their change detection, Liu et al.,~\cite{liu2022cnn}combines a multi-scale CNN-transformer structure to enhance change detection performance in remote sensing images~\cite{liu2022cnn}. Many of these fully supervised change detection methods adopts multi-level representation learning to achieve supervior performance, however, training these supervised models requires pixel level labeling of each image pair, which is time consuming and labor intensive.

\begin{figure*}[t]
\begin{center}
   \includegraphics[width=0.8\linewidth]{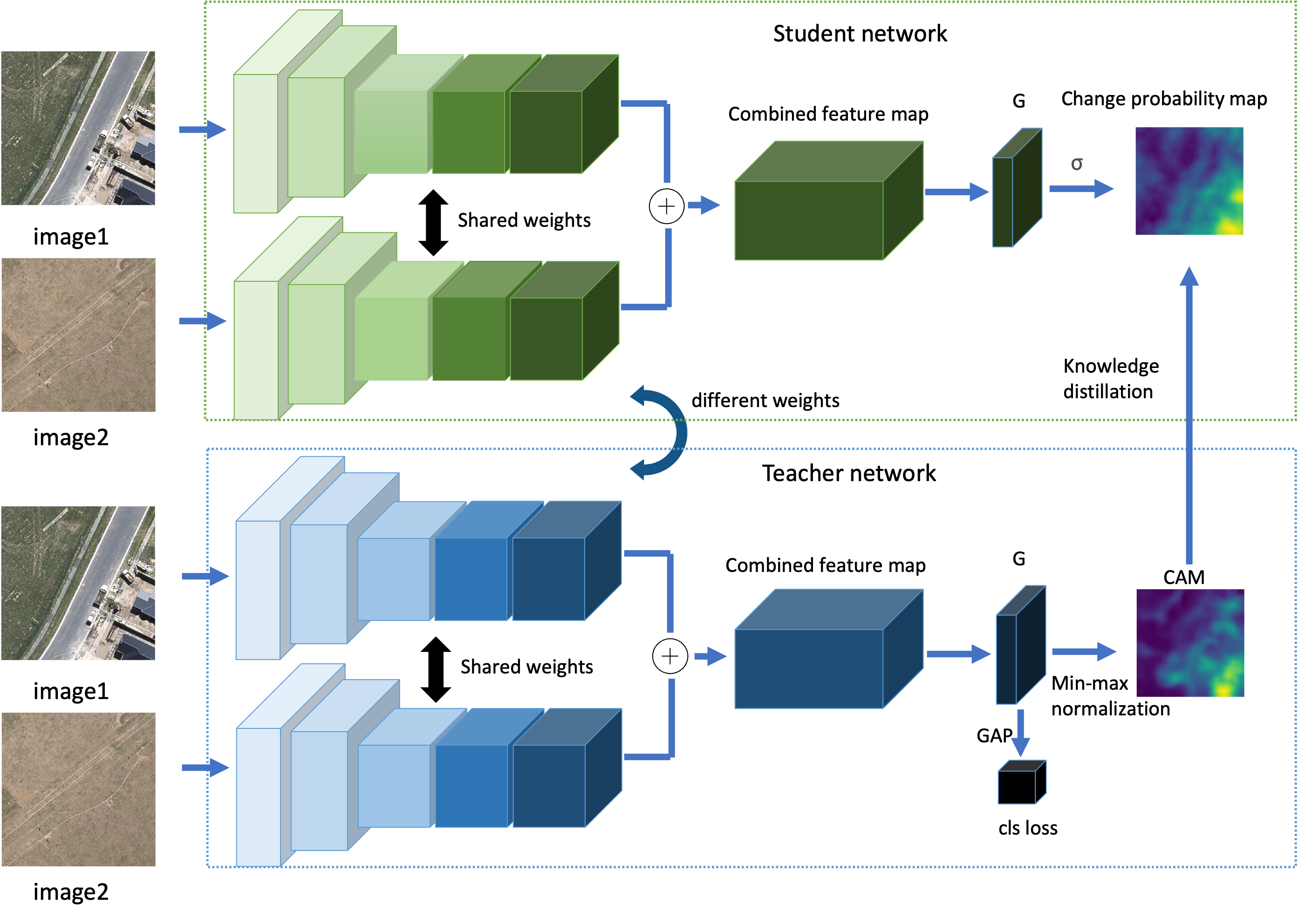}
\end{center}\vspace{-3mm}
   \caption{\small Weakly supervised change detection network with a knowledge distillation framework. Both student and teacher networks are Siamese networks. $\bigoplus$ stands for feature combination methods to combine two time period feature maps, \textit{e.g.}, subtraction, absolute subtraction, and concatenation. Both teacher and student networks are trained concurrently. G stands for one channel feature map, GAP stands for global average pooling, CAM stands for class activation map, $\sigma$ is sigmoid activation function}
\label{fig:2}
\end{figure*}

\subsection{Weakly-supervised semantic segmentation}
Weakly supervised semantic segmentation usually extracts attention map from classification network to serve as pseudo segmentation labels, which are further used to train a segmentation models. The focus of weakly supervised semantic segmentation is to generate high quality attention map\cite{wang2020self}\cite{lee2021reducing}\cite{wei2017object}\cite{zhang2020survey}. For example, Wei et al.\cite{wei2017object} proposed adversarial erasing strategy, which iterately erases the most discriminative region in its attention map in order to find more missing region for the object. Kolesnikov et al.\cite{kolesnikov2016seed} introduce the seed expansion and constraint method which tries to expand the initial seed from attention map to align with the object boundaries. Wang et al. \cite{wang2020self} built a siamese netowrk utilizing the scale invariant property of attention map to make it cover more object regions. Lee et al. \cite{lee2021reducing} modified the last layer of a deep neural network based on information bottleneck theory to mine more non-discriminative region of the object. Wang et al.\cite{chang2020weakly} introduced a novel approach that enhances weakly-supervised semantic segmentation by exploiting sub-category information. Specifically, the method involves clustering image features to generate pseudo sub-category labels within each annotated parent class and constructing a sub-category objective, leading to improved quality of response maps and segmentation results by encouraging the network to focus beyond the most discriminative object parts.
\subsection{Weakly supervised change detection}
Due to the heavy labeling cost of fully supervised change detection models, some researchers have studied the direction to train change detection models with weak labels such as image level labels. Khan et al.\cite{khan2016learning} incorporated conditional random fields in their research. Andermatt et al\cite{andermatt2020weakly} introduced W-CDNet which can be trained with image-level semantic labels for change detection, employing a W-shaped siamese U-net and a Change Segmentation and Classification (CSC) module to create and refine change masks. Meanwhile, Kalita et al.\cite{kalita2021land} employed both principal component analysis and K-means clustering in their methodologies. Wu et al\cite{wu2023fully} leveraged adversarial learning techniques during their model's training phase. Additionally, Huang et al.\cite{huang2023background} introduced an innovative augmentation technique that blends the background areas of image pairs, specifically for weak change detection.
\subsection{Knowledge distillation}
Knowledge distillation, a concept introduced and developed in seminal works\mbox{\cite{hinton2015distilling}}\mbox{\cite{furlanello2018born}}\cite{gou2021knowledge}\cite{gou2021knowledge}, involves training a compact student model to replicate the behavior of a larger, pre-trained teacher model\cite{wang2021knowledge}. This technique has gained prominence for its effectiveness in model compression and facilitating the transfer of knowledge from complex models to simpler ones\cite{alkhulaifi2021knowledge}. In classification tasks, the student model assimilates knowledge by approximating the output distribution of the teacher model, effectively capturing the nuanced relationships learned by the teacher.


Recent advancements have extended the application of knowledge distillation to more complex tasks such as semantic segmentation ~\cite{he2019knowledge}~\cite{liu2019structured}~\cite{qin2021efficient}~\cite{dou2020unpaired}~\cite{ji2022structural}, demonstrating its versatility and potential in various domains. In our study, we adopt the framework of knowledge distillation with a novel focus: applying it to the domain of weakly supervised change detection. Specifically, we aim to distill knowledge from Class Activation Maps (CAMs), leveraging them to guide the student model in an online learning framework. This approach is particularly innovative as it navigates the challenges of weakly supervised settings, where the scarcity of labeled data can impede the learning process.

\section{Proposed Method}

Our proposed method's innovation includes two key components: 1) a knowledge distillation framework for generating more accurate change probability map, and 2) a multiscale sigmoid inference module as post-processing to improve the accuracy of change probability map which will serve as pseudo pixel label to train a change detection model.



\subsection{The Knowledge Distillation Framework}

The knowledge distillation framework consists of two subnetworks, \textit{i.e.}, a Siamese teacher network that can generate Class Activation Maps (CAM)~\cite{zhou2016learning} based on the image-level labels, \textit{i.e.}, change or no change, and a Siamese student network that can generate a fine-grained change probability map via knowledge distillation. Both networks are jointly optimized to enable the student network to highlight potential change areas more accurately than the teacher network. The details of the proposed knowledge distillation framework are shown in Figure~\ref{fig:2}.

\subsubsection{Siamese Teacher Network} 

Siamese teacher network aims to encode the image-level label information and produce CAM to guide the Siamese student network. Given pre-event and post-event image pair \((I_1, I_2)\) where $I_1\in\mathbb{R}^{m\times n\times c}$ and $I_2\in\mathbb{R}^{m\times n\times c}$ and image-level label $y\in\{0,1\}$, each image is passed through the ResNet50~\cite{he2016deep} backbone network to get their last layer of high dimensional feature maps \(F_1\) and \(F_2\), respectively. 
 \(F_1 \bigoplus F_2\) is further passed through another 1$\times$1 convolutional layer and gets the feature map \(G_{\textrm{teacher}}\) with a channel size of 1, where $\bigoplus$ stands for feature combination methods, \textit{e.g.}, concatenation, subtraction, absolute subtraction. In our experiments, we choose the combination that provides the best validation IoU. After that, the CAM can be inferred via min-max normalization:
\begin{equation}
\textrm{CAM}(G_{\textrm{teacher}}) = \frac{ReLU(G_{\textrm{teacher}})}{max(ReLU(G_{\textrm{teacher}}))}.
\label{eq:2}
\end{equation}
Finally, global average pooling (GAP) is applied to the feature map $G_{\textrm{teacher}}$ and a binary cross-entropy loss is employed to train the network, \textit{i.e.},

\begin{align}
\small
L_{\text{cls}} =  &-y \log(\sigma(\textrm{GAP}(G_{\textrm{teacher}})))\\
                  &+ (1-y) \log(1-\sigma(\textrm{GAP}(G_{\textrm{teacher}}))).\nonumber
\end{align}
where $\sigma(\cdot)$ stands for the sigmoid activation function.

Although CAM can provide localization capability for image-level labels, it only highlights the region of the actual change at the coarse level since the goal of the classification network is to classify rather than localize.

\subsubsection{Siamese Student Network}

Siamese student network aims to learn fine-grained change probability map based on the knowledge distilled from the CAM. The Siamese student network shares the same network architecture as the teacher network, however, the weights are trained separately. Similar to the teacher network, we can obtain the feature map \(G_{\textrm{student}}\) based on the same pair of pre-event and post-event image input \((I_1, I_2)\) from the teacher network. Based on that, we apply sigmoid activation to the one-channel feature map \(G_{\textrm{student}}\) as student network's change probability map \( \sigma (G_{\text{student}}) \). Finally, knowledge distillation is conducted by minimizing the Mean Square Error (MSE) between the student change probability map of the student network and the CAM of the teacher network. Specifically, the knowledge distillation loss is given by:

\begin{equation}
L_{\text{kd}} = \| \textrm{CAM}({G_{\text{teacher}}) - \sigma (G_{\text{student}})} \|_2^2.
\end{equation}

\subsubsection{Learning Objective}

The overall learning objective of the knowledge distillation framework can be written as:
\begin{equation}
L =L_{\text{cls}}+ \lambda L_{\text{kd}}, \label{eq:4}
\end{equation}
where $\lambda>0$ is a hyperparameter to control the trade-off between those two terms. Once the model is trained, only the student model is used to infer the change probability map.

\subsection{Multiscale Sigmoid Inference}

In weakly supervised semantic segmentation, multiscale inference has been employed to further boost the performance of CAM. Specifically, it will first resize the image to a set of predefined scales \( S = \{0.5, 1.0, 1.5, 2.0\} \). Then, for each scale, the image is further flipped along the height dimension, doubling the set of image variations. Under the knowledge distillation framework, we adapt the multiscale inference to the Siamese student network and further develop multiscale sigmoid inference to enhance the change probability map of the Siamese student network.

\subsubsection{Multiscale Inference}
We first adapt the multiscale inference to the Siamese student network. Let $I_1^{s}$ and $I_2^{s}$ denote the original images at scale s where \( s \in S \),
$I_1^{sf}$ and $I_2^{sf}$ represent their flipped counterparts.  For each pair of different scales and flip or non-flip images, they are fed into the trained Siamese student network to get the corresponding 1-channel feature map \(G_{\textrm{student}}^s\) and \(G_{\textrm{student}}^{sf}\), respectively. The feature maps of the flipped image pair are then flipped to the original orientation as $G_{\textrm{student}}^{sff}$. After that, all of the feature maps are further resized to the input image size as  $G_{\textrm{student}}^{sr}$ and $G_{\textrm{student}}^{sffr}$. These resized feature map are summed up to get $G_{\textrm{student}}^{\text{sum}}$ which can be given as:
\begin{equation}
G_{\textrm{student}}^{\text{sum}} = \sum_{s \in S} \left(G_{\textrm{student}}^{sr} +  G_{\textrm{student}}^{sffr} )\right), \nonumber
\end{equation}
and \(G_{\textrm{student}}^{\text{sum}}\) is further passed through Equation\ref{eq:2} to obtain the student network's change probability map. One potential issue with the multiscale inference with our student network is that the min-max normalization during in Equation\ref{eq:2} causes a potential mismatch of distribution with the sigmoid activation which is applied during the training of the student network and thus may not get the desired refinement of the change probability map.


\subsubsection{Multiscale Sigmoid Inference}

To resolve the aforementioned issue, we design a new Multiscale Sigmoid Inference (MSI) module to further enhance the change probability map for the Siamese student network. For input $I_1^{s}$ and $I_2^{s}$ at scale s where \( s \in S \), where \( S = \{0.5, 1.0, 1.5, 2.0\} \). Following the similar procedure in multiscale inference, we can obtain $G_{\textrm{student}}^{sr}$ and $G_{\textrm{student}}^{sffr}$. Before summing them up, we first apply
the sigmoid activation function to ensure each scale of feature map can take equally amount of weight and obtain $\sigma (G_{\textrm{student}}^{sr})$ and $\sigma (G_{\textrm{student}}^{sffr})$, respectively. Then, we take their average to obtain the final change probability map, \textit{i.e.},
\begin{align}
M_\textrm{student}=\frac{1}{2|S|} \sum_{s \in S} \left( \sigma (G_{\textrm{student}}^{sr}) \right. \left. + \sigma (G_{\textrm{student}}^{sffr}) \right). \nonumber
\end{align}
The pseudo code of multi-scale sigmoid inference is illustrated in Algorithm 1. Based on the change probability map $M_\textrm{student}$, a background channel with a certain threshold can be applied and the argmax operation can be used to obtain the pseudo pixel-level label for the student network. This pseudo label will be employed to train a separate change detection network for evaluation and test(details are provided in the experiments).

\begin{algorithm}
\caption{Multiscale Sigmoid Inference (MSI)}
\begin{algorithmic}
\State \textbf{Input:} Image pairs ($I_1$, $I_2$), scales $S$, Siamese student network $N_{\text{student}}$
\State \textbf{Output:} Change Probability Map

\Function{MultiscaleSigmoidInference}{$I_1, I_2, S, N_{\text{student}}$}
\For{each $s \in S$}
\State Resize images to scale $s$: $I_1^{s}, I_2^{s}$
\State Flip resized images: $I_1^{sf}, I_2^{sf}$
\For{each pair $(I_1^{s}, I_2^{s}), (I_1^{sf}, I_2^{sf})$}
\State Pass through $N_{\text{student}}$ for pre-logit maps $G_{\text{student}}^{s}, G_{\text{student}}^{sf}$
\State Apply sigmoid activation $\sigma$: $ \sigma(G_{\text{student}}^{s}), \sigma(G_{\text{student}}^{sf})$

\If{pair is flipped}
\State Flip $\sigma(G_{\text{student}}^{sf})$ back to original orientation as $\sigma(G_{\text{student}}^{sff})$
\EndIf
\State Resize $\sigma(G_{\text{student}}^{s}), \sigma(G_{\text{student}}^{sff})$ to input size as $\sigma(G_{\text{student}}^{sr}), \sigma(G_{\text{student}}^{sffr})$
\EndFor
\EndFor
\State \Return $\frac{1}{2|S|} \sum_{s \in S} \left( \sigma (G_{\text{student}}^{sr}) \right. \left. + \sigma (G_{\text{student}}^{sffr}) \right). \nonumber$
\EndFunction
\end{algorithmic}
\end{algorithm}

\section{Experiments}

\subsection{Datasets}

Our research involved conducting experiments on three change detection (CD) datasets. The first dataset is the LEVIR-CD~\cite{chen2020spatial}, which stands for Learning, Vision, and Remote sensing. This is a publicly available large-scale building CD dataset containing 637 pairs of high-resolution (HR) remote sensing (RS) images with a resolution of 0.5 meters and dimensions of 1024 × 1024 pixels. We adhered to the standard dataset split provided for training, validation, and testing. Due to limitations in GPU memory capacity, we divided these images into smaller, non-overlapping patches of 256 × 256 pixels, resulting in 7120, 1024, and 2048 pairs of patches for the training, validation, and test sets, respectively.

The second dataset utilized was the WHU-CD~\cite{ji2018fully}, provided by Wuhan University. This dataset comprises a pair of HR (0.075 meters resolution) aerial images with dimensions of 32,507 × 15,354 pixels, focusing on building change detection. Since the original dataset did not provide a data split solution, we processed the images into non-overlapping patches of 256 × 256 pixels and then randomly divided them into training, validation, and test sets with 5947, 743, and 744 patches, respectively.

In our study, we also utilized the DSIFN-CD dataset~\cite{zhang2020deeply}. This dataset is publicly available and forms part of the Deeply Supervised Image Fusion Network project. It includes a collection of six extensive pairs of high-resolution (HR) satellite images, each with a resolution of 2 meters, sourced from six major cities in China. The dataset is diverse, encompassing a variety of land cover changes like alterations in roads, buildings, croplands, and water bodies. Originally, the DSIFN-CD dataset comprised 3600 training samples, 340 validation samples, and 48 test samples. Each sample measured 512 × 512 pixels in size. To better suit our experimental requirements, we further segmented these images into smaller, non-overlapping patches of 256 × 256 pixels. This process resulted in an increased count of samples, with 14400 training, 1360 validation, and 192 test patches derived from the original dataset allocations. Since the train test ratio of this dataset is highly imbalanced, we reallocated 1638 images from the training set to the test set. This adjustment led to a new distribution of samples for our experiments: 12762 patches for training, 1360 for validation, and 1830 for testing. This restructured dataset allowed for a more balanced and effective evaluation of our change detection methodologies.

\subsection{Setup and Implementation Details}

In our work, we utilize ResNet50~\cite{he2016deep} as the backbone for both the Siamese teacher network and the Siamese student network. Within each Siamese network (teacher or student), the ResNet50 backbones share weights, ensuring a mirrored structure. However, the teacher and student networks do not share weights and are trained jointly with the loss in Eq.~\ref{eq:4}

We train the proposed network with training data from the WHU-CD, DSIFN-CD dataset and LEVIR-CD datasets with image-level labels. The pixel-level labels for the training data are used for evaluating IoU over the change probability map. IoU of the change class is used as a metric to determine the early stop. The weight $\lambda$ for $L_{\text{kd}}$ is set as 10 based on the validation set. 
Note that only the teacher network is trained with the binary cross entropy loss whereas the student network doesn't have classification loss.

The network is trained on an NVIDIA GeForce RTX 3090 GPU with 24GB of VRAM with a batch size of 8 for 20 epochs. The initial learning rate is 0.001 with polynomial learning rate decay. After that, only the student network is kept to obtain the change probability map, which is further refined by multi-scale sigmoid inference to obtain pseudo ground truth. Next, the pseudo pixel-level labels are treated as ground truth to train a separate change detection network with a batch size of 16 for 50 epochs. The initial learning rate is 0.007 with polynomial learning rate decay. 

For the change detection, we employ DeepLabV3+~\cite{chen2018encoder} with Resnet50 for the encoder and modify it as a Siamese network. The image pair are first fed to the encoder to obtain their corresponding high-level and low-level feature maps as $F_1$, $F_2$, $F_{1low}$, and $F_{2low}$. $F_1$-$F_2$ is passed through the same ASPP module from DeepLabV3+ model and its output is further concatenated with $F_{1low}$-$F_{2low}$ to serve as the input to the same decoder from DeepLabV3+ to get the final change mask.

\subsection{Metrics}
To comprehensively evaluate the performance of our model, we employ a suite of metrics: Overall Accuracy (OA), F1-score (F1), Change Class Intersection over Union (IoU), False Positives (FP), False Negatives (FN), and Mean Intersection over Union (Mean IoU).

\begin{itemize}
    \item \textbf{Overall Accuracy (OA)} quantifies the proportion of correctly predicted pixels in the total number of predicted pixels. It is calculated as:
\begin{equation}
OA = \frac{\text{number of correctly predicted pixels}}{\text{total number of pixels}}.
\end{equation}

\item \textbf{F1-score (F1)} serves as the harmonic mean of precision and recall, offering a balance between these two metrics. It is particularly useful for datasets with imbalanced class distributions. The F1-score is formulated as:
\begin{equation}
F1 = 2 \times \frac{\text{precision} \times \text{recall}}{\text{precision} + \text{recall}}.
\end{equation}

\item \textbf{Change Class Intersection over Union (IoU)} measures the overlap between the predicted change pixels and the actual change pixels. It is defined as:
\begin{equation}
cIoU = \frac{\text{TP}}{\text{TP} + \text{FP} + \text{FN}},
\end{equation}
where TP (True Positives) are the correctly predicted change pixels.

\item \textbf{False Positives (FP)} are those instances where non-change pixels are incorrectly identified as changes by the model. 

\item \textbf{False Negatives (FN)} occur when actual change pixels are missed by the model. Both FP and FN are critical for understanding the types of errors made by the model.

\item \textbf{Mean Intersection over Union (Mean IoU)} is an average of the IoU values across all classes, providing a comprehensive view of the model’s performance across different types of changes. It is especially relevant in scenarios involving multiple classes:
\begin{equation}
\text{mIoU} = \frac{1}{N} \sum_{i=1}^{N} IoU_{i},
\end{equation}
where N is the number of classes, in our change detection settings, the number of class is 2. $IoU_i$ is the IoU for the i-th class.

\end{itemize}


\begin{table}[t]
\centering
\caption{Comparison with state-of-the-art methods on WHU-CD test dataset}
\setlength{\tabcolsep}{3.5pt}
\renewcommand{\arraystretch}{1.5} 
\begin{tabular}{ccccccc}
\hline
Method & F1 $\uparrow$ & OA $\uparrow$ & cIoU $\uparrow$ & mIoU $\uparrow$ & FP $\downarrow$ & FN $\downarrow$\\
\hline

FCDNet & 0.645 & 0.937 & 0.193 & 0.564 & 0.491 & 0.317\\
WCDNet & 0.732 & 0.962 & 0.319 & 0.640 & \underline{0.284} & 0.398\\
CAM & \underline{0.797} & \underline{0.966} & \underline{0.441} & \underline{0.703} & 0.356 & \underline{0.203}\\
\hline
Ours & \textbf{0.854} & \textbf{0.977} & \textbf{0.562} & \textbf{0.769} & \textbf{0.245} & \textbf{0.193}\\
\hline
Supervised & 0.944 & 0.992 & 0.807 & 0.899 & 0.069 & 0.124\\
\hline
\end{tabular}
\label{table:1}
\end{table}

\begin{table}  
\centering
\caption{Comparison with state-of-the-art methods on LEVIR-CD test dataset}
\setlength{\tabcolsep}{3.5pt}
\renewcommand{\arraystretch}{1.5} 
\begin{tabular}{ccccccc}
\hline
Method & F1 $\uparrow$ & OA $\uparrow$ & cIoU $\uparrow$ & mIoU $\uparrow$ & FP $\downarrow$ & FN $\downarrow$\\
\hline

FCDNet & 0.551 & 0.888 & 0.088 & 0.487 & 0.585 & 0.328\\
WCDNet & 0.728 & \underline{0.938} & 0.324 & \underline{0.630} & \textbf{0.450} & 0.225\\
CAM & \underline{0.729} & 0.934 & \underline{0.327} & \underline{0.630} & 0.480 & \underline{0.193}\\
\hline
Ours & \textbf{0.749} & \textbf{0.939} & \textbf{0.361} & \textbf{0.649} & \underline{0.464} & \textbf{0.174}\\
\hline
Supervised & 0.922 & 0.985 & 0.742 & 0.863 & 0.115 & 0.142\\
\hline
\end{tabular}
\label{table:2}
\end{table}



\begin{table}[htbp]
\centering
\caption{Comparison with state-of-the-art methods on DSIFN-CD test dataset}
\setlength{\tabcolsep}{3.5pt}
\renewcommand{\arraystretch}{1.5} 
\begin{tabular}{ccccccc}
\hline
Method & F1 $\uparrow$ & OA $\uparrow$ & cIoU $\uparrow$ & mIoU $\uparrow$ & FP $\downarrow$ & FN $\downarrow$\\
\hline

FCDNet & 0.278 & 0.355 & 0.345 & 0.184 & 0.654 & \textbf{0.001}\\
WCDNet & \underline{0.704} & \underline{0.754} & 0.412 & \underline{0.557} & \textbf{0.186} & 0.402\\
CAM & 0.664 & 0.666 & \underline{0.471} & 0.498 & 0.461 & \underline{0.068}\\
\hline
Ours & \textbf{0.757} & \textbf{0.775} & \textbf{0.529} & \textbf{0.614} & \underline{0.287} & 0.183\\
\hline
Supervised & 0.929 & 0.935 & 0.831 & 0.868 & 0.112 & 0.057\\
\hline
\end{tabular}
\label{table:3}
\end{table}

\subsection{Comparison with State-of-the-Art}
There is only limited weakly change detection literature available. We adopt FCDNet\cite{wu2023fully} and WCDNet\cite{andermatt2020weakly} as two of our baselines since they provide their code available online. FCDNet\cite{wu2023fully} applied adversarial learning during the training of their change detection network with image-level labels. WCDNet\cite{andermatt2020weakly} developed a change segmentation and classification module with image-level labels only, where the change segmentation mask is learned inside the module. CRF-RNN module is then applied to further refine the change segmentation mask. We also consider a CAM baseline to compare with our proposed network. CAM baseline uses ResNet50 as the backbone of the Siamese teacher network and employs multiscale inference to refine CAM and obtain the pseudo pixel-level label. The same Siamese DeepLabV3+~\cite{chen2018encoder} model is used to pursue change detection. We also compared the fully supervised change detection method. The fully supervised change detection model is the same siamese DeepLabV3+~\cite{chen2018encoder} but trained with the original dataset's pixel-level labels. The quantitative comparison results on the test data of the LEVIR-CD, WHU-CD, and DSIFN-CD datasets are shown in Table~\ref{table:1}, Table~\ref{table:2}, and Table~\ref{table:3}. We can observe that our proposed method consistently outperforms state-of-the-art models on all three datasets. In particular, the proposed method outperforms CAM which only leverages the Siamese teacher network, this demonstrate the effectiveness of knowledge distillation for producing the change probability map in the student network.

\begin{figure}
  \centering
  \includegraphics[width=0.5\textwidth]{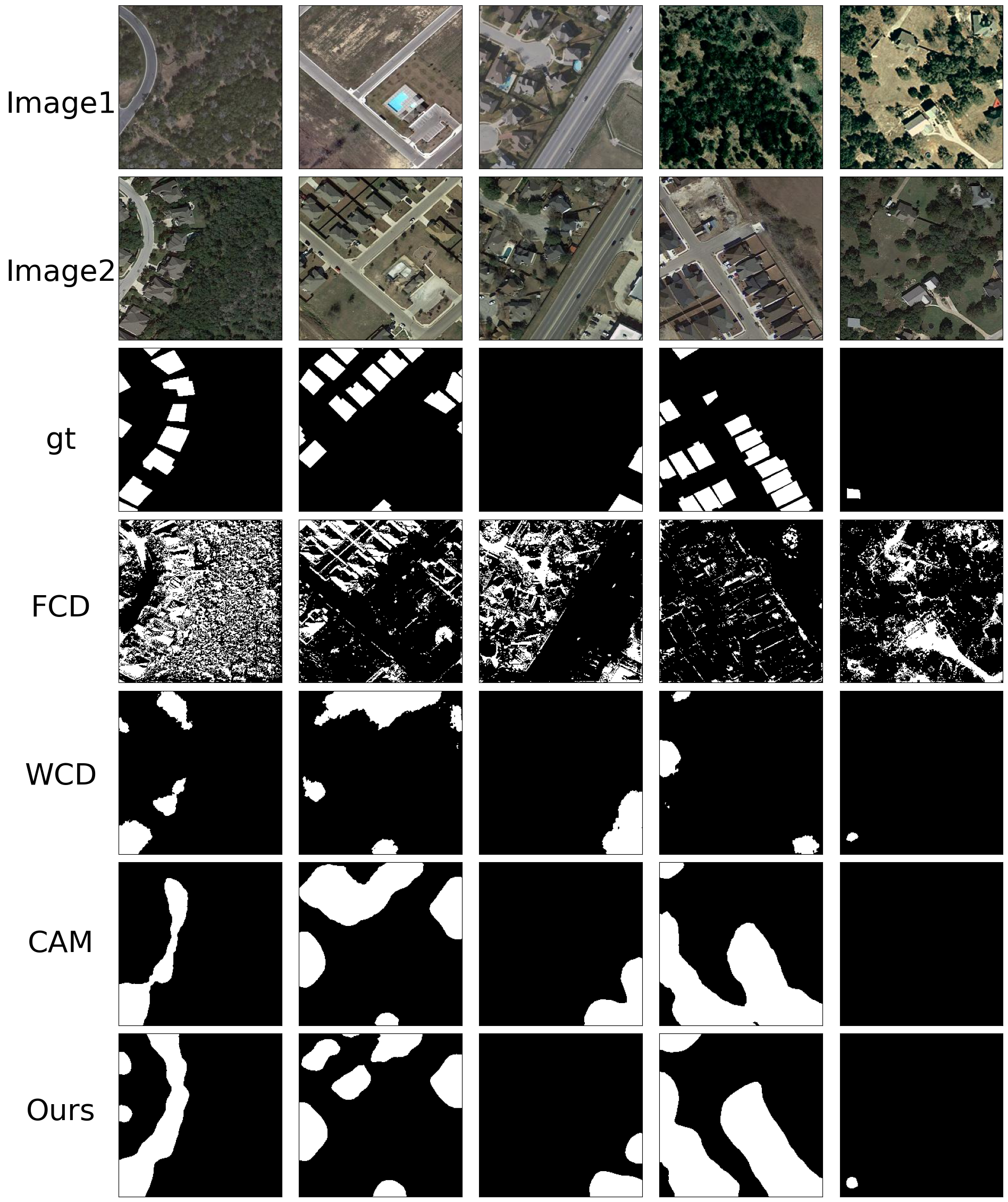}
  \caption{LEVIR-CD test dataset performance comparison.}
  \label{fig:image3}
\end{figure}

Some visual comparisons on the test set of LEVIR-CD and WHU-CD datasets are shown in Figure~\ref{fig:image3} and Figure~\ref{fig:image4} respectively. Visual comparison on DSIFN-CD test dataset is shown in supplemental material. We can also observe that our proposed method can produce more accurate pixel-level change labels compared to all three baselines.


\begin{figure}[h!]
  \centering
  \includegraphics[width=0.5\textwidth]{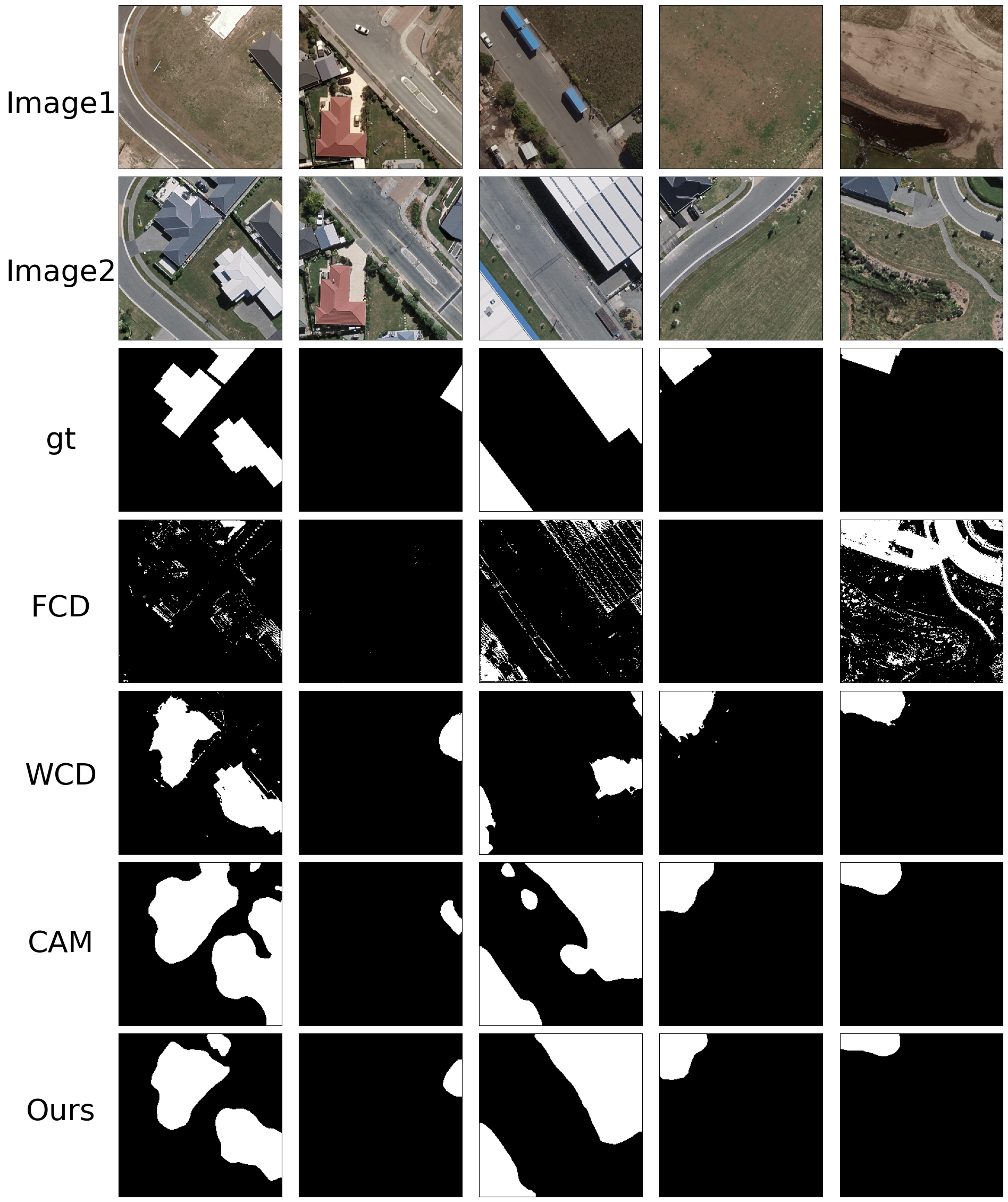}
  \caption{WHU-CD test dataset performance comparison.}
  \label{fig:image4}
\end{figure}

\subsection{Ablation Study}

Based on the WHU-CD dataset training data, we perform an ablation study to justify the effectiveness of each component of our proposed work. We first compare the change probability map from the student network and CAM from the teacher model. The student network's change probability map can improve the teacher network's CAM IoU from 35.1$\%$ to 47.7$\%$ with 12.6$\%$ improvement. We also compare the result of the student network's change probability map under the multi-scale inference method (MI) and multi-scale sigmoid inference method (MSI). We observe that MI may slightly reduce its IoU from 47.7$\%$ to 47.3$\%$. But with MSI, we can achieve 52.7$\%$, with a 5$\%$ improvement over the change probability map for the student network with knowledge distillation. 

We also visualize the learned CAM from the teacher network (column (d)), the change probability map from the student network (column (e)), and the change probability map from the student network with multi-scale sigmoid inference (MSI) (column (f))in Fig.~\ref{fig:4}. Our results show that the change probability map from the student network with MSI can better highlight the change regions.

\begin{table}[h]
  \centering
    \setlength{\tabcolsep}{10pt} 
    \caption{Ablation study on WHU-CD dataset, where MI denotes multi-scale inference, MSI stands for multi-scale sigmoid inference.}
  \begin{tabular}{|c|c|c|c|c|c| }
    \hline
    \textbf{Teacher} & \textbf{Student} & \textbf{MI} & \textbf{MSI}  &\textbf{IoU}\\
    \hline
     $\checkmark$ &  &  & &  0.351 \\
     & $\checkmark$ &  & &  0.477 \\
     & $\checkmark$ & $\checkmark$ & &  0.473 \\
     & $\checkmark$ &  & $\checkmark$&  0.527 \\
    \hline
  \end{tabular}

  \label{table:4}
\end{table}

\begin{figure} [!h]
    \centering
    \setlength{\tabcolsep}{0.5pt} 

    \begin{tabular}{*{6}{c}}

        \begin{subfigure}{0.14\linewidth}
            \includegraphics[width=\linewidth]{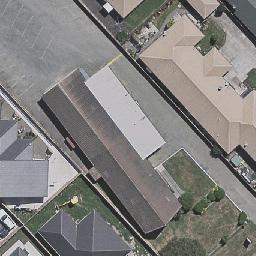}
        \end{subfigure} &
        \begin{subfigure}{0.14\linewidth}
            \includegraphics[width=\linewidth]{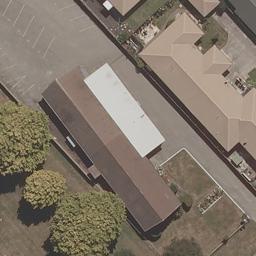}
        \end{subfigure} &
        \begin{subfigure}{0.14\linewidth}
            \includegraphics[width=\linewidth]{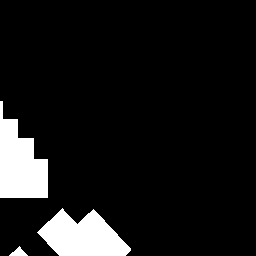}
        \end{subfigure} &
        \begin{subfigure}{0.14\linewidth}
            \includegraphics[width=\linewidth]{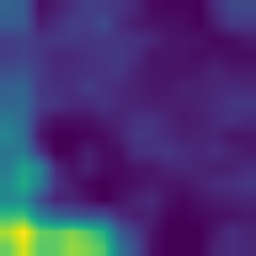}
        \end{subfigure} &
        \begin{subfigure}{0.14\linewidth}
            \includegraphics[width=\linewidth]{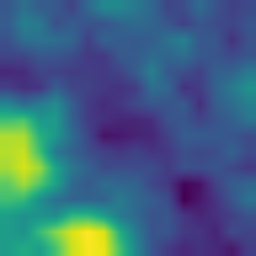}
        \end{subfigure} &
        \begin{subfigure}{0.14\linewidth}
            \includegraphics[width=\linewidth]{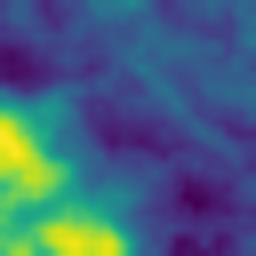}
        \end{subfigure} \\[0em]


        \begin{subfigure}{0.14\linewidth}
            \includegraphics[width=\linewidth]{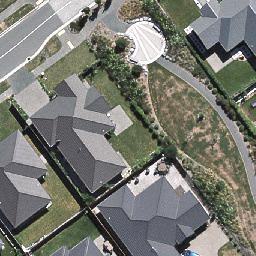}
        \end{subfigure} &
        \begin{subfigure}{0.14\linewidth}
            \includegraphics[width=\linewidth]{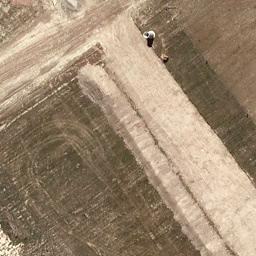}
        \end{subfigure} &
        \begin{subfigure}{0.14\linewidth}
            \includegraphics[width=\linewidth]{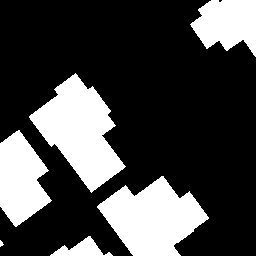}
        \end{subfigure} &
        \begin{subfigure}{0.14\linewidth}
            \includegraphics[width=\linewidth]{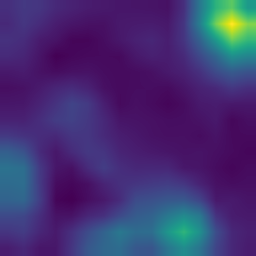}
        \end{subfigure} &
        \begin{subfigure}{0.14\linewidth}
            \includegraphics[width=\linewidth]{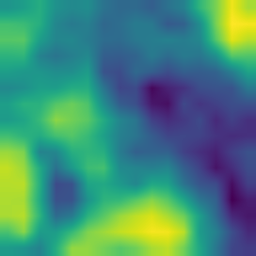}
        \end{subfigure} &
        \begin{subfigure}{0.14\linewidth}
            \includegraphics[width=\linewidth]{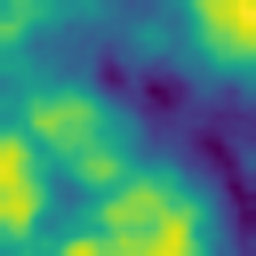}
        \end{subfigure} \\[0em]


        \begin{subfigure}{0.14\linewidth}
            \includegraphics[width=\linewidth]{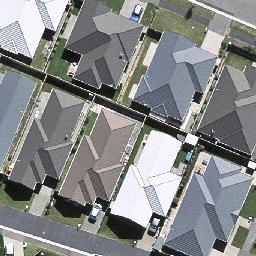}
            \caption*{(a)}
        \end{subfigure} &
        \begin{subfigure}{0.14\linewidth}
            \includegraphics[width=\linewidth]{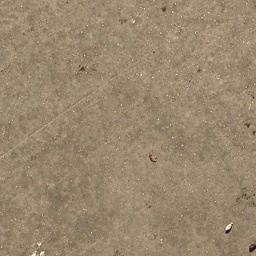}
            \caption*{(b)}
        \end{subfigure} &
        \begin{subfigure}{0.14\linewidth}
            \includegraphics[width=\linewidth]{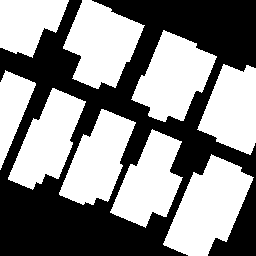}
            \caption*{(c)}
        \end{subfigure} &
        \begin{subfigure}{0.14\linewidth}
            \includegraphics[width=\linewidth]{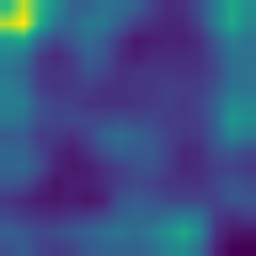}
            \caption*{(d)}
        \end{subfigure} &
        \begin{subfigure}{0.14\linewidth}
            \includegraphics[width=\linewidth]{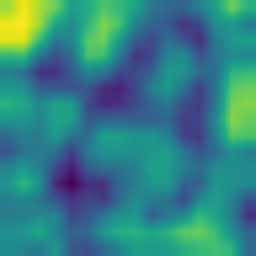}
            \caption*{(e)}
        \end{subfigure} &
        \begin{subfigure}{0.14\linewidth}
            \includegraphics[width=\linewidth]{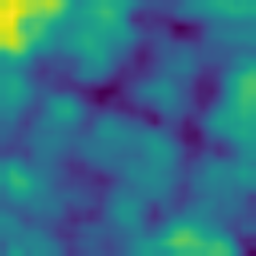}
            \caption*{(f)}
        \end{subfigure} \\

    \end{tabular}
    \caption{\small Visualization of change probability map on WHU-CD dataset, where column(a) denote pre-event images, column(b) represents post-event images, column(c) represents ground truth, column(d) represents CAM from teacher model, column(e) represents change probability map from student model, column(f) stands for change probability from student model with MSI. }
    \label{fig:4}
\end{figure}

\section{Conclusion}

In this paper, we developed a novel weakly supervised change detection method based on a knowledge distillation framework and multi-scale sigmoid inference module. Extensive experiments on three public datasets, \textit{i.e.}, WHU-CD, LEVIR-CD, and DSIFN-CD, justified the effectiveness of our proposed model.





\clearpage

\bibliographystyle{named}
\bibliography{ijcai24}

\begin{thebibliography}{}

\bibitem[\protect\citeauthoryear{Alkhulaifi \bgroup \em et al.\egroup }{2021}]{alkhulaifi2021knowledge}
Abdolmaged Alkhulaifi, Fahad Alsahli, and Irfan Ahmad.
\newblock Knowledge distillation in deep learning and its applications.
\newblock {\em PeerJ Computer Science}, 7:e474, 2021.

\bibitem[\protect\citeauthoryear{Andermatt and Timofte}{2020}]{andermatt2020weakly}
Philipp Andermatt and Radu Timofte.
\newblock A weakly supervised convolutional network for change segmentation and classification.
\newblock In {\em Proceedings of the Asian Conference on Computer Vision}, 2020.

\bibitem[\protect\citeauthoryear{Bouziani \bgroup \em et al.\egroup }{2010}]{bouziani2010automatic}
Mourad Bouziani, Kalifa Go{\"\i}ta, and Dong-Chen He.
\newblock Automatic change detection of buildings in urban environment from very high spatial resolution images using existing geodatabase and prior knowledge.
\newblock {\em ISPRS Journal of Photogrammetry and Remote Sensing}, 65(1):143--153, 2010.

\bibitem[\protect\citeauthoryear{Chang \bgroup \em et al.\egroup }{2020}]{chang2020weakly}
Yu-Ting Chang, Qiaosong Wang, Wei-Chih Hung, Robinson Piramuthu, Yi-Hsuan Tsai, and Ming-Hsuan Yang.
\newblock Weakly-supervised semantic segmentation via sub-category exploration.
\newblock In {\em Proceedings of the IEEE/CVF Conference on Computer Vision and Pattern Recognition}, pages 8991--9000, 2020.

\bibitem[\protect\citeauthoryear{Chen and Shi}{2020}]{chen2020spatial}
Hao Chen and Zhenwei Shi.
\newblock A spatial-temporal attention-based method and a new dataset for remote sensing image change detection.
\newblock {\em Remote Sensing}, 12(10):1662, 2020.

\bibitem[\protect\citeauthoryear{Chen \bgroup \em et al.\egroup }{2018}]{chen2018encoder}
Liang-Chieh Chen, Yukun Zhu, George Papandreou, Florian Schroff, and Hartwig Adam.
\newblock Encoder-decoder with atrous separable convolution for semantic image segmentation.
\newblock In {\em Proceedings of the European conference on computer vision (ECCV)}, pages 801--818, 2018.

\bibitem[\protect\citeauthoryear{Chen \bgroup \em et al.\egroup }{2021}]{chen2021remote}
Hao Chen, Zipeng Qi, and Zhenwei Shi.
\newblock Remote sensing image change detection with transformers.
\newblock {\em IEEE Transactions on Geoscience and Remote Sensing}, 60:1--14, 2021.

\bibitem[\protect\citeauthoryear{Daudt \bgroup \em et al.\egroup }{2018}]{daudt2018fully}
Rodrigo~Caye Daudt, Bertr Le~Saux, and Alexandre Boulch.
\newblock Fully convolutional siamese networks for change detection.
\newblock In {\em 2018 25th IEEE International Conference on Image Processing (ICIP)}, pages 4063--4067. IEEE, 2018.

\bibitem[\protect\citeauthoryear{Dou \bgroup \em et al.\egroup }{2020}]{dou2020unpaired}
Qi~Dou, Quande Liu, Pheng~Ann Heng, and Ben Glocker.
\newblock Unpaired multi-modal segmentation via knowledge distillation.
\newblock {\em IEEE transactions on medical imaging}, 39(7):2415--2425, 2020.

\bibitem[\protect\citeauthoryear{Furlanello \bgroup \em et al.\egroup }{2018}]{furlanello2018born}
Tommaso Furlanello, Zachary Lipton, Michael Tschannen, Laurent Itti, and Anima Anandkumar.
\newblock Born again neural networks.
\newblock In {\em International Conference on Machine Learning}, pages 1607--1616. PMLR, 2018.

\bibitem[\protect\citeauthoryear{Gou \bgroup \em et al.\egroup }{2021}]{gou2021knowledge}
Jianping Gou, Baosheng Yu, Stephen~J Maybank, and Dacheng Tao.
\newblock Knowledge distillation: A survey.
\newblock {\em International Journal of Computer Vision}, 129:1789--1819, 2021.

\bibitem[\protect\citeauthoryear{Goyette \bgroup \em et al.\egroup }{2012}]{goyette2012changedetection}
Nil Goyette, Pierre-Marc Jodoin, Fatih Porikli, Janusz Konrad, and Prakash Ishwar.
\newblock Changedetection. net: A new change detection benchmark dataset.
\newblock In {\em 2012 IEEE computer society conference on computer vision and pattern recognition workshops}, pages 1--8. IEEE, 2012.

\bibitem[\protect\citeauthoryear{He \bgroup \em et al.\egroup }{2016}]{he2016deep}
Kaiming He, Xiangyu Zhang, Shaoqing Ren, and Jian Sun.
\newblock Deep residual learning for image recognition.
\newblock In {\em Proceedings of the IEEE conference on computer vision and pattern recognition}, pages 770--778, 2016.

\bibitem[\protect\citeauthoryear{He \bgroup \em et al.\egroup }{2019}]{he2019knowledge}
Tong He, Chunhua Shen, Zhi Tian, Dong Gong, Changming Sun, and Youliang Yan.
\newblock Knowledge adaptation for efficient semantic segmentation.
\newblock In {\em Proceedings of the IEEE/CVF Conference on Computer Vision and Pattern Recognition}, pages 578--587, 2019.

\bibitem[\protect\citeauthoryear{Hinton \bgroup \em et al.\egroup }{2015}]{hinton2015distilling}
Geoffrey Hinton, Oriol Vinyals, and Jeff Dean.
\newblock Distilling the knowledge in a neural network.
\newblock {\em arXiv preprint arXiv:1503.02531}, 2015.

\bibitem[\protect\citeauthoryear{Huang \bgroup \em et al.\egroup }{2023}]{huang2023background}
Rui Huang, Ruofei Wang, Qing Guo, Jieda Wei, Yuxiang Zhang, Wei Fan, and Yang Liu.
\newblock Background-mixed augmentation for weakly supervised change detection.
\newblock In {\em Proceedings of the AAAI Conference on Artificial Intelligence}, volume~37, pages 7919--7927, 2023.

\bibitem[\protect\citeauthoryear{Ji \bgroup \em et al.\egroup }{2018}]{ji2018fully}
Shunping Ji, Shiqing Wei, and Meng Lu.
\newblock Fully convolutional networks for multisource building extraction from an open aerial and satellite imagery data set.
\newblock {\em IEEE Transactions on geoscience and remote sensing}, 57(1):574--586, 2018.

\bibitem[\protect\citeauthoryear{Ji \bgroup \em et al.\egroup }{2022}]{ji2022structural}
Deyi Ji, Haoran Wang, Mingyuan Tao, Jianqiang Huang, Xian-Sheng Hua, and Hongtao Lu.
\newblock Structural and statistical texture knowledge distillation for semantic segmentation.
\newblock In {\em Proceedings of the IEEE/CVF Conference on Computer Vision and Pattern Recognition}, pages 16876--16885, 2022.

\bibitem[\protect\citeauthoryear{Jiang \bgroup \em et al.\egroup }{2020}]{jiang2020pga}
Huiwei Jiang, Xiangyun Hu, Kun Li, Jinming Zhang, Jinqi Gong, and Mi~Zhang.
\newblock Pga-siamnet: Pyramid feature-based attention-guided siamese network for remote sensing orthoimagery building change detection.
\newblock {\em Remote Sensing}, 12(3):484, 2020.

\bibitem[\protect\citeauthoryear{Jiang \bgroup \em et al.\egroup }{2022}]{jiang2022survey}
Huiwei Jiang, Min Peng, Yuanjun Zhong, Haofeng Xie, Zemin Hao, Jingming Lin, Xiaoli Ma, and Xiangyun Hu.
\newblock A survey on deep learning-based change detection from high-resolution remote sensing images.
\newblock {\em Remote Sensing}, 14(7):1552, 2022.

\bibitem[\protect\citeauthoryear{Kalita \bgroup \em et al.\egroup }{2021}]{kalita2021land}
Indrajit Kalita, Savvas Karatsiolis, and Andreas Kamilaris.
\newblock Land use change detection using deep siamese neural networks and weakly supervised learning.
\newblock In {\em Computer Analysis of Images and Patterns: 19th International Conference, CAIP 2021, Virtual Event, September 28--30, 2021, Proceedings, Part II 19}, pages 24--35. Springer, 2021.

\bibitem[\protect\citeauthoryear{Khan \bgroup \em et al.\egroup }{2016}]{khan2016learning}
Salman~H Khan, Xuming He, Fatih Porikli, Mohammed Bennamoun, Ferdous Sohel, and Roberto Togneri.
\newblock Learning deep structured network for weakly supervised change detection.
\newblock {\em arXiv preprint arXiv:1606.02009}, 2016.

\bibitem[\protect\citeauthoryear{Khelifi and Mignotte}{2020}]{khelifi2020deep}
Lazhar Khelifi and Max Mignotte.
\newblock Deep learning for change detection in remote sensing images: Comprehensive review and meta-analysis.
\newblock {\em Ieee Access}, 8:126385--126400, 2020.

\bibitem[\protect\citeauthoryear{Kolesnikov and Lampert}{2016}]{kolesnikov2016seed}
Alexander Kolesnikov and Christoph~H Lampert.
\newblock Seed, expand and constrain: Three principles for weakly-supervised image segmentation.
\newblock In {\em Computer Vision--ECCV 2016: 14th European Conference, Amsterdam, The Netherlands, October 11--14, 2016, Proceedings, Part IV 14}, pages 695--711. Springer, 2016.

\bibitem[\protect\citeauthoryear{Lee \bgroup \em et al.\egroup }{2021a}]{lee2021local}
Haeyun Lee, Kyungsu Lee, Jun~Hee Kim, Younghwan Na, Juhum Park, Jihwan~P Choi, and Jae~Youn Hwang.
\newblock Local similarity siamese network for urban land change detection on remote sensing images.
\newblock {\em IEEE Journal of Selected Topics in Applied Earth Observations and Remote Sensing}, 14:4139--4149, 2021.

\bibitem[\protect\citeauthoryear{Lee \bgroup \em et al.\egroup }{2021b}]{lee2021reducing}
Jungbeom Lee, Jooyoung Choi, Jisoo Mok, and Sungroh Yoon.
\newblock Reducing information bottleneck for weakly supervised semantic segmentation.
\newblock {\em Advances in Neural Information Processing Systems}, 34:27408--27421, 2021.

\bibitem[\protect\citeauthoryear{Liu \bgroup \em et al.\egroup }{2019}]{liu2019structured}
Yifan Liu, Ke~Chen, Chris Liu, Zengchang Qin, Zhenbo Luo, and Jingdong Wang.
\newblock Structured knowledge distillation for semantic segmentation.
\newblock In {\em Proceedings of the IEEE/CVF conference on computer vision and pattern recognition}, pages 2604--2613, 2019.

\bibitem[\protect\citeauthoryear{Liu \bgroup \em et al.\egroup }{2020}]{liu2020building}
Yi~Liu, Chao Pang, Zongqian Zhan, Xiaomeng Zhang, and Xue Yang.
\newblock Building change detection for remote sensing images using a dual-task constrained deep siamese convolutional network model.
\newblock {\em IEEE Geoscience and Remote Sensing Letters}, 18(5):811--815, 2020.

\bibitem[\protect\citeauthoryear{Liu \bgroup \em et al.\egroup }{2022}]{liu2022cnn}
Mengxi Liu, Zhuoqun Chai, Haojun Deng, and Rong Liu.
\newblock A cnn-transformer network with multiscale context aggregation for fine-grained cropland change detection.
\newblock {\em IEEE Journal of Selected Topics in Applied Earth Observations and Remote Sensing}, 15:4297--4306, 2022.

\bibitem[\protect\citeauthoryear{Qin \bgroup \em et al.\egroup }{2021}]{qin2021efficient}
Dian Qin, Jia-Jun Bu, Zhe Liu, Xin Shen, Sheng Zhou, Jing-Jun Gu, Zhi-Hua Wang, Lei Wu, and Hui-Fen Dai.
\newblock Efficient medical image segmentation based on knowledge distillation.
\newblock {\em IEEE Transactions on Medical Imaging}, 40(12):3820--3831, 2021.

\bibitem[\protect\citeauthoryear{Shi \bgroup \em et al.\egroup }{2020}]{shi2020change}
Wenzhong Shi, Min Zhang, Rui Zhang, Shanxiong Chen, and Zhao Zhan.
\newblock Change detection based on artificial intelligence: State-of-the-art and challenges.
\newblock {\em Remote Sensing}, 12(10):1688, 2020.

\bibitem[\protect\citeauthoryear{Simonyan and Zisserman}{2014}]{simonyan2014very}
Karen Simonyan and Andrew Zisserman.
\newblock Very deep convolutional networks for large-scale image recognition.
\newblock {\em arXiv preprint arXiv:1409.1556}, 2014.

\bibitem[\protect\citeauthoryear{Sublime and Kalinicheva}{2019}]{sublime2019automatic}
J{\'e}r{\'e}mie Sublime and Ekaterina Kalinicheva.
\newblock Automatic post-disaster damage mapping using deep-learning techniques for change detection: Case study of the tohoku tsunami.
\newblock {\em Remote Sensing}, 11(9):1123, 2019.

\bibitem[\protect\citeauthoryear{Wang and Yoon}{2021}]{wang2021knowledge}
Lin Wang and Kuk-Jin Yoon.
\newblock Knowledge distillation and student-teacher learning for visual intelligence: A review and new outlooks.
\newblock {\em IEEE transactions on pattern analysis and machine intelligence}, 44(6):3048--3068, 2021.

\bibitem[\protect\citeauthoryear{Wang \bgroup \em et al.\egroup }{2020}]{wang2020self}
Yude Wang, Jie Zhang, Meina Kan, Shiguang Shan, and Xilin Chen.
\newblock Self-supervised equivariant attention mechanism for weakly supervised semantic segmentation.
\newblock In {\em Proceedings of the IEEE/CVF conference on computer vision and pattern recognition}, pages 12275--12284, 2020.

\bibitem[\protect\citeauthoryear{Wei \bgroup \em et al.\egroup }{2017}]{wei2017object}
Yunchao Wei, Jiashi Feng, Xiaodan Liang, Ming-Ming Cheng, Yao Zhao, and Shuicheng Yan.
\newblock Object region mining with adversarial erasing: A simple classification to semantic segmentation approach.
\newblock In {\em Proceedings of the IEEE conference on computer vision and pattern recognition}, pages 1568--1576, 2017.

\bibitem[\protect\citeauthoryear{Wu \bgroup \em et al.\egroup }{2023}]{wu2023fully}
Chen Wu, Bo~Du, and Liangpei Zhang.
\newblock Fully convolutional change detection framework with generative adversarial network for unsupervised, weakly supervised and regional supervised change detection.
\newblock {\em IEEE Transactions on Pattern Analysis and Machine Intelligence}, 2023.

\bibitem[\protect\citeauthoryear{Ye \bgroup \em et al.\egroup }{2021}]{ye2021near}
Su~Ye, John Rogan, Zhe Zhu, and J~Ronald Eastman.
\newblock A near-real-time approach for monitoring forest disturbance using landsat time series: Stochastic continuous change detection.
\newblock {\em Remote Sensing of Environment}, 252:112167, 2021.

\bibitem[\protect\citeauthoryear{Zhang \bgroup \em et al.\egroup }{2020a}]{zhang2020deeply}
Chenxiao Zhang, Peng Yue, Deodato Tapete, Liangcun Jiang, Boyi Shangguan, Li~Huang, and Guangchao Liu.
\newblock A deeply supervised image fusion network for change detection in high resolution bi-temporal remote sensing images.
\newblock {\em ISPRS Journal of Photogrammetry and Remote Sensing}, 166:183--200, 2020.

\bibitem[\protect\citeauthoryear{Zhang \bgroup \em et al.\egroup }{2020b}]{zhang2020survey}
Man Zhang, Yong Zhou, Jiaqi Zhao, Yiyun Man, Bing Liu, and Rui Yao.
\newblock A survey of semi-and weakly supervised semantic segmentation of images.
\newblock {\em Artificial Intelligence Review}, 53:4259--4288, 2020.

\bibitem[\protect\citeauthoryear{Zhang \bgroup \em et al.\egroup }{2022}]{zhang2022swinsunet}
Cui Zhang, Liejun Wang, Shuli Cheng, and Yongming Li.
\newblock Swinsunet: Pure transformer network for remote sensing image change detection.
\newblock {\em IEEE Transactions on Geoscience and Remote Sensing}, 60:1--13, 2022.

\bibitem[\protect\citeauthoryear{Zheng \bgroup \em et al.\egroup }{2021}]{zheng2021building}
Zhuo Zheng, Yanfei Zhong, Junjue Wang, Ailong Ma, and Liangpei Zhang.
\newblock Building damage assessment for rapid disaster response with a deep object-based semantic change detection framework: From natural disasters to man-made disasters.
\newblock {\em Remote Sensing of Environment}, 265:112636, 2021.

\bibitem[\protect\citeauthoryear{Zhou \bgroup \em et al.\egroup }{2016}]{zhou2016learning}
Bolei Zhou, Aditya Khosla, Agata Lapedriza, Aude Oliva, and Antonio Torralba.
\newblock Learning deep features for discriminative localization.
\newblock In {\em Proceedings of the IEEE conference on computer vision and pattern recognition}, pages 2921--2929, 2016.

\end{thebibliography}

\end{document}